# A first step towards automated species recognition from camera trap images of mammals using AI in a European temperate forest

Mateusz Choiński[1], Mateusz Rogowski[1], Piotr Tynecki[1], Dries P.J. Kuijper[2], Marcin Churski[2], Jakub W. Bubnicki[2]

[1]Computer Science Faculty of Bialystok University of Technology, Poland
[2]Mammal Research Institute, Polish Academy of Sciences, Poland
{m.choinski, p.tynecki}@doktoranci.pb.edu.pl

**Abstract.** Camera traps are used worldwide to monitor wildlife. Despite the increasing availability of Deep Learning (DL) models, the effective usage of this technology to support wildlife monitoring is limited. This is mainly due to the complexity of DL technology and high computing requirements. This paper presents the implementation of the light-weight and state-of-the-art YOLOv5 architecture for automated labeling of camera trap images of mammals in the Białowieża Forest (BF), Poland. The camera trapping data were organized and harmonized using TRAPPER software, an open source application for managing large-scale wildlife monitoring projects. The proposed image recognition pipeline achieved an average accuracy of 85% F1-score in the identification of the 12 most commonly occurring medium-size and large mammal species in BF using a limited set of training and testing data (a total 2659 images with animals).

Based on the preliminary results, we concluded that the YOLOv5 object detection and classification model is a promising light-weight DL solution after the adoption of transfer learning technique. It can be efficiently plugged in via an API into existing web-based camera trapping data processing platforms such as e.g. TRAPPER system. Since TRAPPER is already used to manage and classify (manually) camera trapping datasets by many research groups in Europe, the implementation of AI-based automated species classification may significantly speed up the data processing workflow and thus better support data-driven wildlife monitoring and conservation. Moreover, YOLOv5 developers perform better performance on edge devices which may open a new chapter in animal population monitoring in real time directly from camera trap devices.

## 1   Introduction

To conserve and manage diverse mammalian communities in a way that their population status is secured and conflict with humans is minimized, requires in the first place comprehensive data-derived knowledge of their status. There is a growing awareness that standard wildlife monitoring methods are not effective and difficult to scale-up (e.g. snow-tracking or hunting-bag data). Therefore, numerous new initiatives to monitor mammals using camera traps are currently being developed across Europe, collectively generating massive amounts of pictures and videos. However a large amount of available data is not effectively exploited, mainly because of the human time needed to mine the data from collected raw multimedia files.

   Camera trapping has already proved to be one of the most important technologies in wildlife conservation and ecological research [1-5]. Rapid developments in the application of AI speed up the transformation in that area and contribute to the fact that most of the recorded material will be automatically classified in the future [6-8, 11-13]. However, despite the increasing availability of deep learning models for object recognition [14, 15, 17, 18], the effective usage of this technology to support wildlife monitoring is limited, mainly because of the complexity of DL technology, the lack of end-to-end pipelines, and high computing requirements.

In this study we present the preliminary results of applying the new AI standalone model for both object detection and species-level classification of camera trapping data from a European temperate lowland forest, the Białowieża Forest, Poland. Our model was built using an extremely fast, lightweight and flexible deep learning architecture based on recently published YOLOv5 (pre-trained YOLOv5l) [9] and trained on 2659 labeled images accessed via the API built in TRAPPER [10]. To the best of our knowledge, this is the first YOLOv5 implementation for automated mammal species recognition using camera trap images.

## 2 Materials and Methods

### 2.1 Dataset preparation and preprocessing

As the main data source in our research we used species-classified images originating from camera trapping projects from Białowieża Primeval Forest stored in TRAPPER [10]. This consisted of 2659 images. Bounding boxes were manually added to all images with animals present to determine the exact position of each individual. When multiple individuals were present in the image, several bounding boxes were created for each individual. Example annotations and images are shown in Fig 3. We did not use empty images in our dataset.

We have 12 classes in our dataset, 11 species of animals and 1 class "Other" which represents birds and small rodents. Images in our dataset were of various sizes. Most of them are 12 Mpixel high-resolution images. The distribution of the sizes can be found in Fig 1. In our dataset we identified 11 common species occurring in Białowieża Primeval Forest.

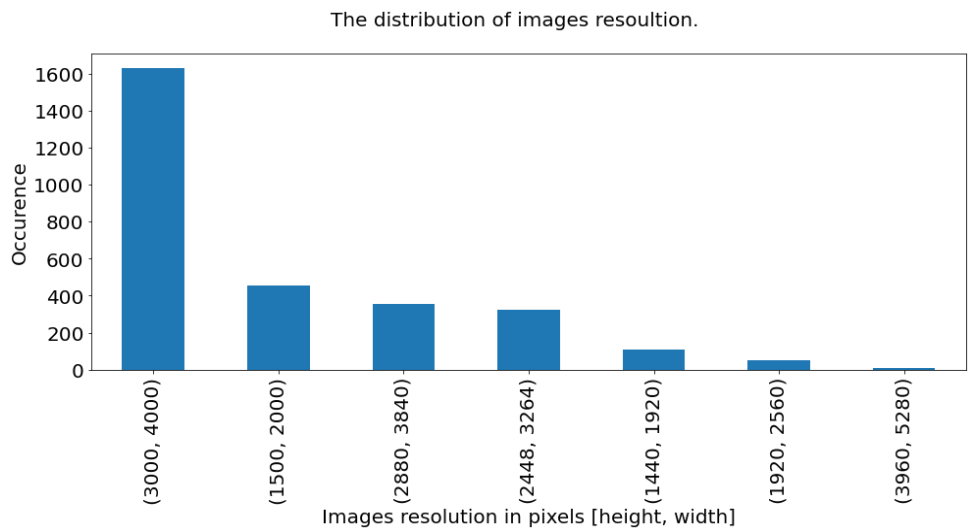

**Fig 1.** The distribution of the image sizes (pixels) in the dataset.

Observations are not balanced across these species and we have species with larger and smaller support of samples (Fig 2). Example annotations and images are shown in Fig 3.

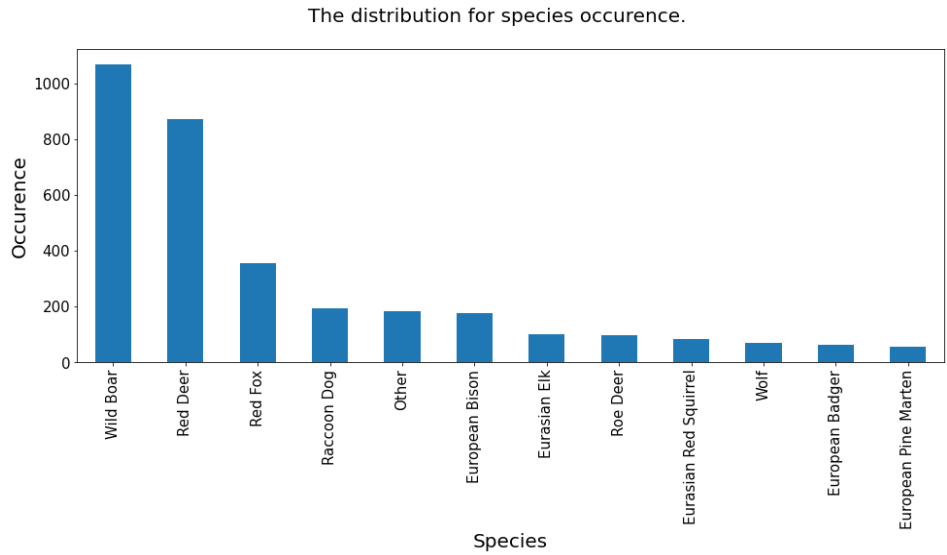

**Fig 2.** The distribution of species in a dataset.

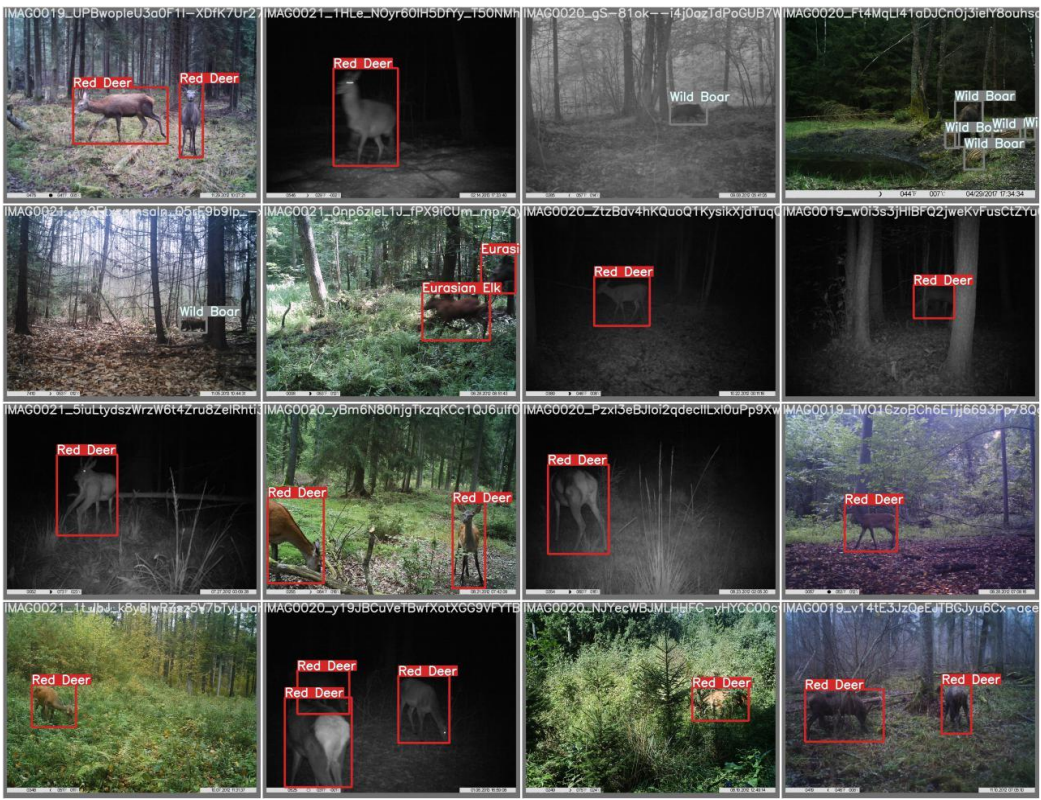

**Fig 3.** Example of images and annotations in TRAPPER platform.

In our experiments we decided to perform cross-validation. This is the process of splitting the dataset into *k* equal size parts and process *k* training runs. For each training run, one split of the data was used for validation, and the rest was used for training the network. Images to each split were selected randomly with stratification on - each separate data split had the same proportions of the observed species.

**2.2 Deep Learning architecture**

The following data pipeline was adopted:
1. The dataset was downloaded from TRAPPER using the dedicated API.
2. Species with less than 40 camera trap images were excluded.
3. The format of the image annotations was adapted to the required input format of the YOLOv5 architecture.
4. Model testing in 5-fold cross validation.

YOLOv5 is the Deep Learning-based architecture we selected for this research. It achieves state of the art results in the object detection field. In comparison to other Deep Learning architectures, YOLOv5 is simple and reliable. It needs much less computational power than other architectures while keeping comparable results [14, 15] and it is at the same time much faster than other networks (Fig 4). YOLOv5 strongly utilizes the architecture of YOLOv4 [18]. The encoder used in YOLOv5 is CSPDarknet [18]. Along with Path Aggregation Network [17] (PANet) they make up the whole network architecture. In comparison to the YOLOv4, activation functions were modified (Leaky ReLU and Hardswish activations were replaced with SiLU [19] activation function).

Selecting the YOLOv5 architecture for this research was motivated by several reasons:
1. The network is currently state of the art in fast objects detections field
2. The architecture is lightweight, this allows us to train the model using small computational resources and keep it cheap
3. Small size of the model may allow it to be used in mobile devices (i.e. camera traps)

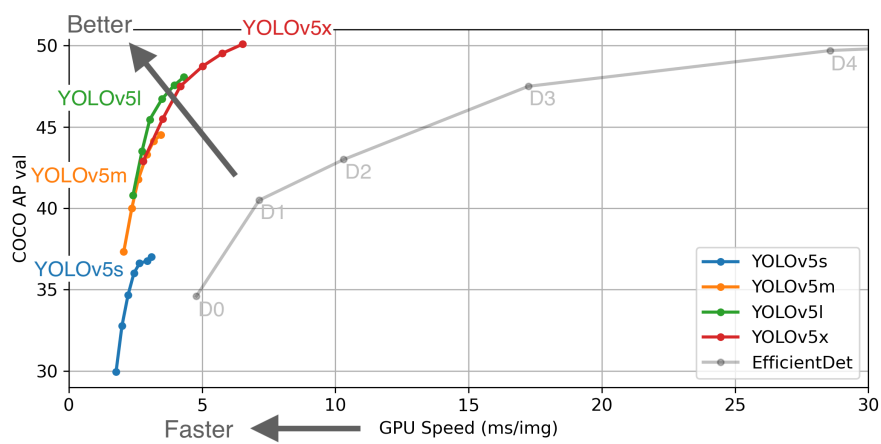

**Fig 4.** Comparisons between YOLOv5 models and EfficientDet.
Published within author permission.

## 2.3 Model training process

The first stage of model training was the hyper-parameter tuning. For that purpose we have used evolutionary hyper-parameter tuning methods from YOLOv5 on the training and validation data. This has given us more optimal parameters for our dataset. In the next step, we have trained the model using the optimal hyper-parameters and it started from an already trained YOLOv5l model checkpoint.

Using a pre-trained model is a common technique in computer vision called Transfer Learning [20]. It speeds up the training process and keeps the generalization on the high level. During our experiments we have observed that the optimal number of epochs is 60, after that there are negligible changes in the model. Fig 5. shows how the YOLOv5 loss functions changed during training, the results are shown on one of the training cross-validation splits.

YOLOv5 loss function is a sum of three smaller loss functions:
- Bounding box regression Loss - penalty for wrong anchor box detection, Mean Squared Error calculated based on predicted box location (x, y, h, w),
- Classification Loss - Cross Entropy calculated for object classification,
- Objectness Loss - Mean Squared Error calculated for Objectness-Confidence Score (estimation if the anchor box contains an object).

Below plots (Fig 5 and Fig 6) shows that after 51 epochs there is a minimal change in the loss functions as well as in F1-score.

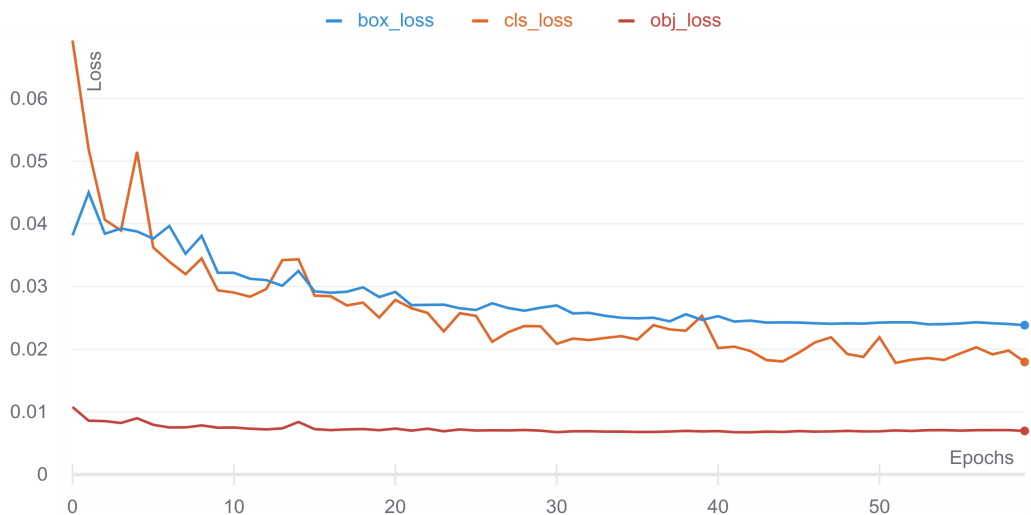

**Fig 5.** Bounding box regression, classification and objectness loss changes during model training on the 1-st split from cross-validation.

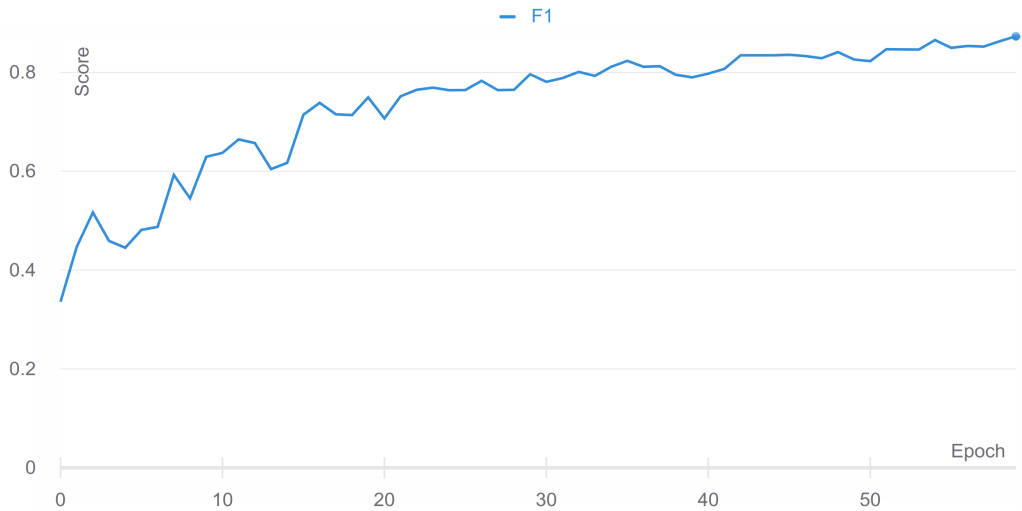

**Fig 6.** F1-score changes during model training on the 1-st split from cross-validation.

The main software used in the training of the models was Python 3.8, with PyTorch 1.7, CUDA 11.2 and Jupyter Notebook. All AI model iterations as well as the final one were trained using cloud computing on Microsoft Azure. We executed the experiment on a virtual machine with preinstalled Linux software and a single NVIDIA Tesla K80 GPU. The training time for a single training run (60 epochs) on the train data takes 2h (8.5h for the whole cross-validation).

## 3   Results and Discussion

The preliminary results using a limited amount of training data, showe on average 85% F1-score in identification of 12 most commonly occuring medium-size and large mammal species in BF (including "other" class). The results presented here are the averaged results from all of the 5 validation splits from the cross-validation process. Table 1. shows the results of that combined evaluation. From the data we can see that the detection precision for species cat down with decreasing sample size. Furthermore, for example, "roe deer" class has the lowest F1-score (0.58) because of the low abundance of this species in our data-set. As a result this species is often not being detected by our model, as shown in Fig 7. Future experiments should address this issue.

Another insight from Fig 7. is that "roe deer" class is often misclassified as "red deer" (15%). These species are similar in size and coloration and the reason for those mistakes might be caused by the low number of the targets - "red deer" occurs 872 times where there are only 97 instances of "roe deer" in the dataset. To address this issue, we suggest to re-run these analyses on larger samples of classified images with bounding boxes.

Table 1. The combined results of the 5-fold cross-validation process.

| Class (Species) | Targets | F1 | P | R | mAP@.5 | mAP@.5:.95 |
|---|---|---|---|---|---|---|
| all | 3143 | **0.85** | 0.88 | 0.82 | 0.88 | 0.66 |
| Wild Boar | 1070 | **0.89** | 0.92 | 0.86 | 0.91 | 0.68 |
| Red Deer | 872 | **0.86** | 0.88 | 0.85 | 0.89 | 0.68 |
| Red Fox | 356 | **0.94** | 0.93 | 0.94 | 0.97 | 0.75 |
| Raccoon Dog | 193 | **0.94** | 0.93 | 0.95 | 0.95 | 0.71 |
| European Bison | 176 | **0.81** | 0.89 | 0.76 | 0.85 | 0.68 |
| Eurasian Elk | 103 | **0.85** | 0.89 | 0.81 | 0.89 | 0.76 |
| Roe Deer | 97 | **0.58** | 0.67 | 0.53 | 0.61 | 0.47 |
| Eurasian Red Squirrel | 84 | **0.89** | 0.93 | 0.84 | 0.91 | 0.59 |
| Wolf | 71 | **0.87** | 0.89 | 0.85 | 0.91 | 0.73 |
| European Badger | 63 | **0.89** | 0.93 | 0.86 | 0.94 | 0.70 |
| European Pine Marten | 58 | **0.76** | 0.83 | 0.72 | 0.80 | 0.54 |

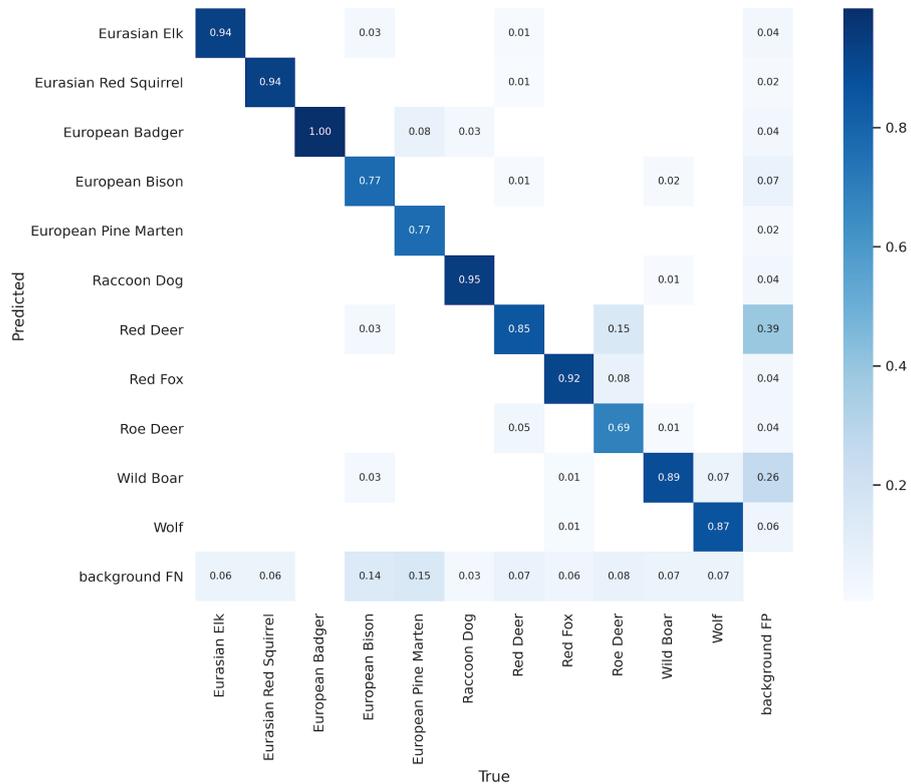

**Fig 7.** Confusion matrix of the predictions on the test data from the 1-st cross-validation split.

### 3.1 Incorrect classification examples

We are aware of the disadvantages of our classification model. Three most common registered issues are: misclassifying animals with trees (Fig 8), classifying animals with incorrect classes and not detecting animals at all (Fig 9).

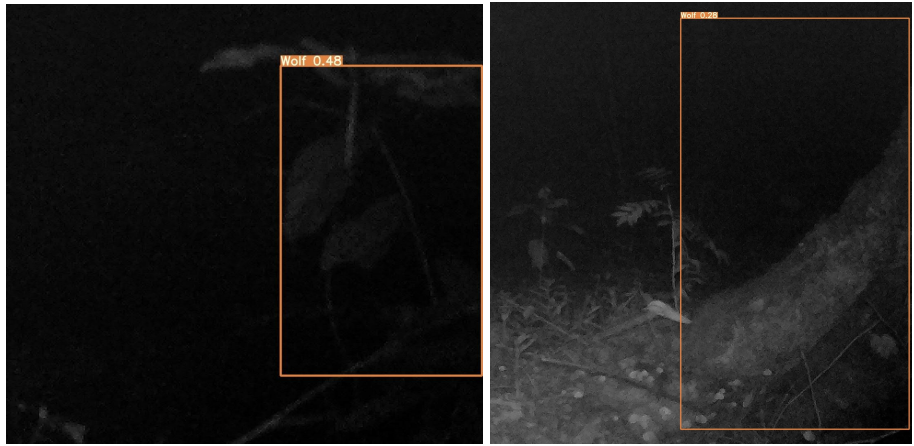

**Fig 8.** Example of trees misclassified as animals.

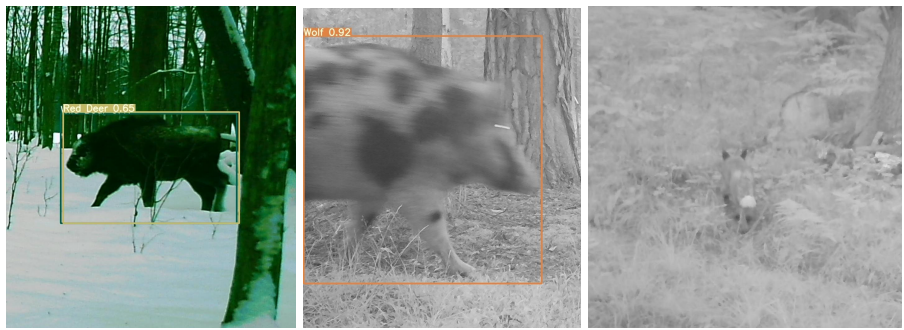

**Fig 9.** Incorrect species classification. European bison is classified as red deer and wild boar as wolf. The last-right image is an example of an unrecognized red fox sample.

### 3.4 Correct classification examples

Tested solution works with expected accuracy level with partially visible animals on images. Fig 10. shows correct detection and classification of a red fox that has only half of its body visible in the frame. Next examples (Fig 11) proves that the model is able to detect animals that are blended with the background.

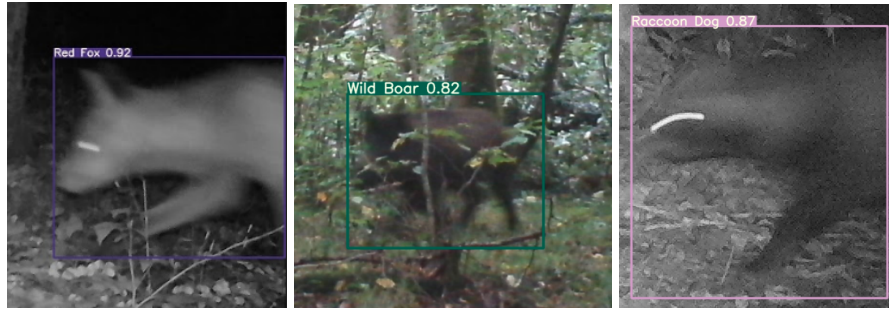

**Fig 10.** First-left image presents correct detection of partially visible animals, the center and last, correct detection of barely visible animals.

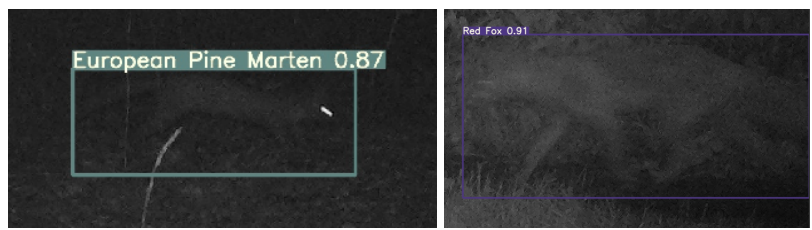

**Fig 11**. Correct detection of animals blended with the background.

## 4   Conclusions

The achieved preliminary results using a limited amount of training data, on average 85% F1-score in identification of 12 most commonly occuring medium-sized and large mammal species in BF (including "other" class). These promising results show a large potential in using YOLOv5 deep learning architecture to train AI model for automatic species recognition. Moreover, it can be directly incorporated into the camera trapping data processing workflows. Extended camera trap images dataset would allow for higher metrics and better detection of species from the Białowieża Forest.

As a future step we see that including additional training data sets from multiple European forest areas would greatly improve generalization of the Deep Learning model. However, this work suggests there is room for improvement in case of the selected YOLOv5 network architecture and the inference pipeline. The encoder network may be changed for deeper structure to be able to extract more specific features from the images. Another common practice to improve the results is usage of the Test Time Augmentation (TTA) on the inference process.

Pre-trained YOLOv5l model is fast enough, accurate and lightweight to be deployed on a well functional Edge AI computing platform. That advantage opens a new chapter for species classification in real time using camera-traps in the field.

**Data availability**

Jupyter Notebook which implements that experiment pipeline includes TRAPPER integration and YOLOv5 usage for species recognition. It was released as Open Source code and is available on GitLab: https://gitlab.com/oscf/trapper-species-classifier

**Conflict of Interest**

The authors declare that they have no conflicts of interest.

**Ethical approval**

All applicable international, national, and/or institutional guidelines for the care and use of animals were followed.

**Acknowledgment**

This work was supported by Microsoft AI for Earth research grant funded by Microsoft.